\pdfoutput=1

\documentclass[11pt]{article}

\usepackage[final]{acl}

\usepackage{times}
\usepackage{latexsym}
\usepackage{tabularx}
\usepackage{booktabs}
\usepackage{makecell}

\usepackage[T1]{fontenc}

\usepackage[utf8]{inputenc}

\usepackage{microtype}

\usepackage{inconsolata}

\usepackage{graphicx}
\usepackage{amsmath}

\title{Reading with Intent - Neutralizing Intent}

\author{Benjamin Reichman, Adar Avsian, Larry Heck \\
  AI Virtual Assistant (AVA) Lab \\
Georgia Institute of Technology \\
  \texttt{\{bzr,aavsian3,larryheck\}@gatech.edu}
  }

\begin{document}
\maketitle
\begin{abstract}
Queries to large language models (LLMs) can be divided into two parts: the instruction/question and the accompanying context. The context for retrieval-augmented generation (RAG) systems in most benchmarks comes from Wikipedia or Wikipedia-like texts which are written in a neutral and factual tone. However, when RAG systems retrieve internet-based content, they encounter text with diverse tones and linguistic styles, introducing challenges for downstream tasks. The Reading with Intent task addresses this issue by evaluating how varying tones in context passages affect model performance. Building on prior work that focused on sarcasm, we extend this paradigm by constructing a dataset where context passages are transformed to $11$ distinct emotions using a better synthetic data generation approach. Using this dataset, we train an emotion translation model to systematically adapt passages to specified emotional tones. The human evaluation shows that the LLM fine-tuned to become the emotion-translator benefited from the synthetically generated data. Finally, the emotion-translator is used in the Reading with Intent task to transform the passages to a neutral tone. By neutralizing the passages, it mitigates the challenges posed by sarcastic passages and improves overall results on this task by about $3\%$.
\end{abstract}

\section{Introduction}
Over the past few years, large language models (LLMs) have vastly expanded in their scope of use, from answering questions and generating code to supporting academic research. Despite their capabilities, LLMs have shortcomings, including a tendency to generate hallucinated content—confidently providing incorrect or fabricated information \cite{chatgpt_hallucinations}. This limitation arises from LLMs having a finite number of parameters that constrain how much knowledge they can learn during pretraining. Furthermore, knowledge is inherently a long-tail problem, making it infeasible for LLMs to memorize all the information necessary to answer every possible query \cite{longtail_knowledge}. This challenge is compounded by the knowledge cutoff date in pretraining, which prevents LLMs from accessing information published after that point. Together, these factors highlight the need for augmenting LLMs with external knowledge sources.

Retrieval-augmented generation (RAG) addresses these limitations by integrating information retrieval with LLMs \cite{rag}. This provides the LLM with relevant facts and passages based on the input query, augmenting it with external knowledge beyond its pretrained parameters. By providing context-specific information, RAG helps the LLM generate more accurate responses and reduces the occurrence of hallucinated content.

In most benchmarks, the retrieval corpus for RAG systems often consists of Wikipedia or Wikipedia-like text, characterized by a neutral, matter-of-fact tone. However, RAG systems deployed with the internet as their retrieval corpus encounter texts with vastly different styles and tones. While Wikipedia adheres to a consistent neutrality, internet texts may embody a range of emotions and linguistic tropes, such as sarcasm, irony, happiness, or excitement. These variations pose significant challenges for LLMs when processing the retrieved context, potentially leading to incorrect, harmful, or toxic outputs in past RAG deployments \cite{googleaioverview,googleaioverview2}.

To address the variability in emotions and linguistic tropes in written text, the Reading with Intent task was introduced by \cite{rwi}. While this work provided a foundation, it had certain limitations. The authors focused on a single linguistic tope and addressed the issue with a lightweight prompting approach.

In this work, we expand the Reading with Intent task and make the following contributions:
\begin{enumerate}
    \item Synthetically generated and analyzed a new Reading with Intent dataset encompassing $11$ distinct emotions.
    \item Develop an emotion translation model capable of adapting text to specified emotional tones.
    \item Evaluate the emotion-translator and the underlying dataset.
    \item Apply the emotion-translator to the Reading with Intent task, demonstrating its impact on task performance.
\end{enumerate}

\begin{figure*}[htb!]
\centering
\includegraphics[width=\textwidth]{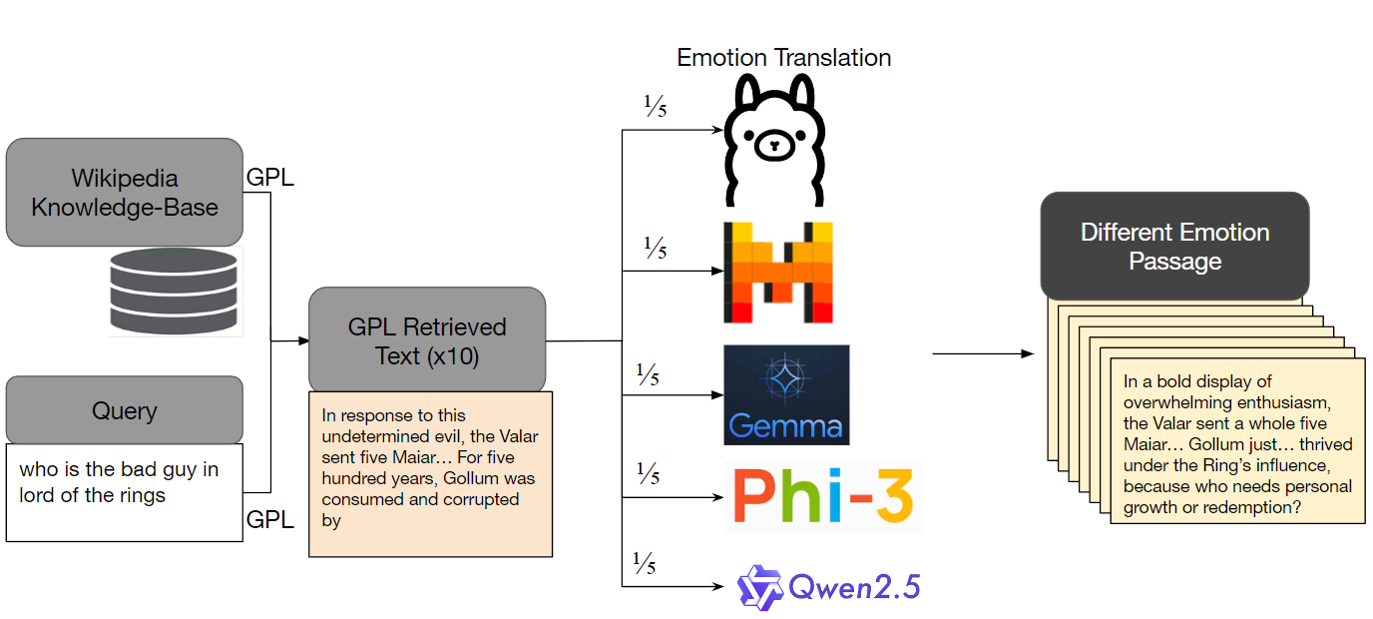}
\caption{Synthetic data generation process.}
\label{fig:dataset_creation}
\end{figure*}

\section{Related Works}
\textbf{Sentiment Analysis:} The task of classifying text based on the emotions it conveys has long been a focus of NLP research. Numerous methods have been proposed to identify emotions in text \cite{prabowo2009sentiment,medhat2014sentiment,wadawadagi2020sentiment,wankhade2022survey}. However, research on reading comprehension tasks involving texts with diverse and heterogeneous emotional tones remains relatively limited. Addressing this gap is critical, as emotional nuances can impact how information in text is interpreted and understood.

\textbf{Style Transfer:} Previous work in translating emotions use style-transfer approaches. One work, for example, involved translating text into a different language to strip its original style, followed by back-translation using an encoder-decoder model with a style-specific decoder \cite{prabhumoye-etal-2018-style}. Another approach developed an augmented zero-shot learning approach, prompting LLMs with examples of multiple style-transfer operations and then asking them to perform a novel style-transfer task not present in the examples \cite{reif-etal-2022-recipe}. Style-transfer approaches for emotion translation has been used in quite a few past approaches \cite{li-etal-2019-domain,qi-etal-2021-mind,Shen2017StyleTF,Yang2018UnsupervisedTS,mir-etal-2019-evaluating}. However, these approaches typically focus on binary or coarse-grained emotions, such as ``positive'' and ``negative'' rather than a broader range of emotions. This paper takes a different approach, emphasizing data-centric methodologies. By synthetically generating a bi-text corpus for an expanded set of emotions, we enable a direct translation approach. This work captures a wider spectrum of emotions than previous style-transfer methods.

\textbf{Sarcasm Detection:} Previous works on sarcasm detection have explored a variety of methodologies. One approach uses convolutional neural networks (CNNs) to extract text features such as sentiment, emotion, and personality, which are then aggregated for overall sarcasm classification \cite{Poria2016ADL}. Another approach employs graph learning to produce sarcasm classifications \cite{Lou2021AffectiveDG}. A further approach leverages commonsense knowledge repositories, such as COMET, to detect sarcasm by reasoning about implicit contextual cues, as demonstrated by \cite{commonsensesarcasm}. These methods are often trained on datasets like SARC and iSarcasm, which provide annotated examples of sarcastic text \cite{sarc,isarcasm}. While these approaches are effective at sarcasm detection in isolation, they are less focused on the challenges of integrating sarcasm detection into downstream tasks, such as reading comprehension. This work builds on these foundations by addressing how sarcasm impacts the interpretability of retrieved passages in Reading with Intent tasks.

\textbf{Sarcasm Generation: } There have been a few prior works on sarcasm generation. One approach employs logical representations to transform sentences into sarcastic versions \cite{oprea-etal-2021-chandler}. Another method reverses the sentiment polarity (valence) of the input sentence and uses the commonsense reasoning framework COMET to generate sarcastic context that aligns with the transformed statement \cite{chakrabarty-etal-2020-r}.

\textbf{Reading Sarcasm: } Less research has been done on integrating sarcasm detection into the reading comprehension task. As LLMs increasingly interact with general internet text rather than carefully curated, Wikipedia-like text, the ability to interpret linguistic tropes like sarcasm becomes essential. Without this capability, models risk misinterpreting sarcastic text and producing harmful or toxic outputs \cite{googleaioverview,googleaioverview2}. Such outputs often stem from a failure to recognize text that, to a human reader, would clearly be intended as jest. The Reading with Intent task \cite{rwi} addresses this challenge by introducing a dataset specifically designed to study how LLMs handle sarcasm in retrieved passages. Building on this foundation, our work leverages emotion translation to improve the readability of sarcastic text for LLMs, enabling more accurate and context-aware processing.

\section{Dataset Creation}

An LLM query can typically be divided into two parts: the object and the context. The object of the query is the question or instruction provided to the LLM—the ``raison d'être'' of the query. The context comprises supplementary text that aids the LLM in producing a more accurate and relevant response to the object of the query. The goal of the created dataset is to systematically alter the emotional tone of the context. 

To create such a dataset, a task that employs the query-context paradigm was needed. Open-domain question answering (QA) was selected due to its reliance on external context and reading comprehension to generate accurate responses. Since our focus is on the context rather than the queries themselves, we utilized the pre-existing Natural Questions dataset, a widely-used open-domain QA benchmark as our base dataset \cite{nq}.


After selecting the QA dataset, the next step was to select context passages for each query. A state-of-the-art open-source retrieval algorithm, GPL, was used \cite{gpl}. Each query in the NQ dataset was embedded using GPL's query encoder and each passage in the associated Wikipedia retrieval corpus was embedded using the passage encoder. The top-10 passages selected for each query were retrieved using maximum inner product search. These retrieved passages form the corpus of contexts in our dataset.

The final step involves modifying the emotions of the passages. Figure \ref{fig:dataset_creation} illustrates the process by which each passage was transformed into 11 distinct emotions or linguistic tropes: anger, condescension, disgust, envy, excitement, fear, happiness, humor, sadness, sarcasm, and surprise. To ensure diversity and mitigate biases from any single model, each passage was randomly assigned to one of five LLMs for each emotion: Llama 3, Qwen 2.5, Phi-3, Gemma, and Mistral-7B. These models were chosen for their varied architectures and training datasets, which allowed for a broader representation of each emotion. 

Each LLM was provided with a specialized prompt tailored to elicit the specified emotional tone or linguistic style. The generated outputs were qualitatively reviewed for consistency, fluency, and alignment with the target emotion. Prompts were iteratively refined based on these reviews to ensure high-quality transformations across all emotions and linguistic tropes. This multi-LLM approach enhanced the robustness and diversity of the resulting dataset.

\begin{figure*}[h!]
\centering
\includegraphics[width=\textwidth]{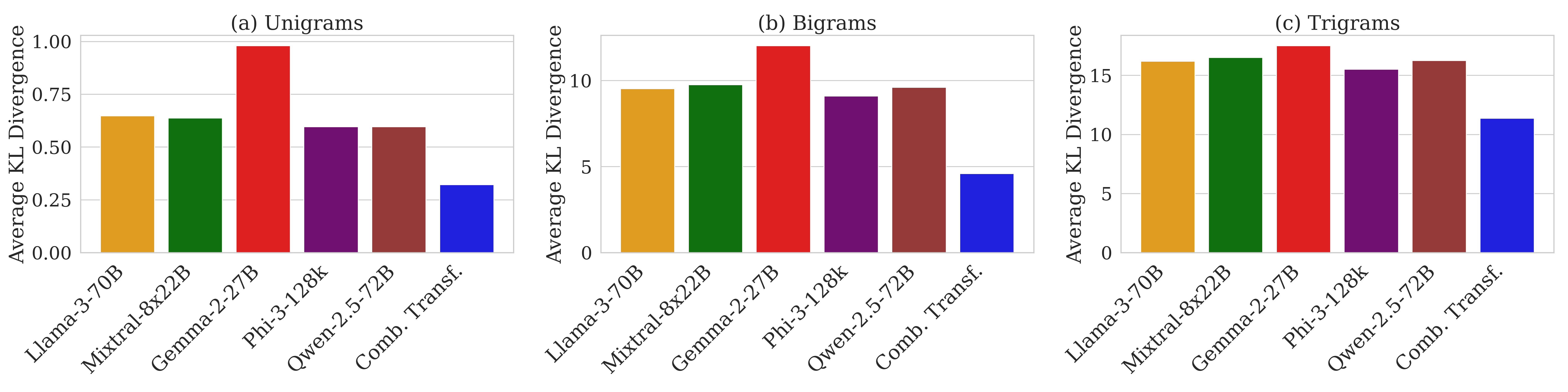}
\caption{The KL-Divergences between the unigram, bigrams, and trigrams of the original and synthetic datasets.}
\label{fig:kl_divergence}
\end{figure*}

\begin{figure}[h!]
\centering
\includegraphics[width=\columnwidth]{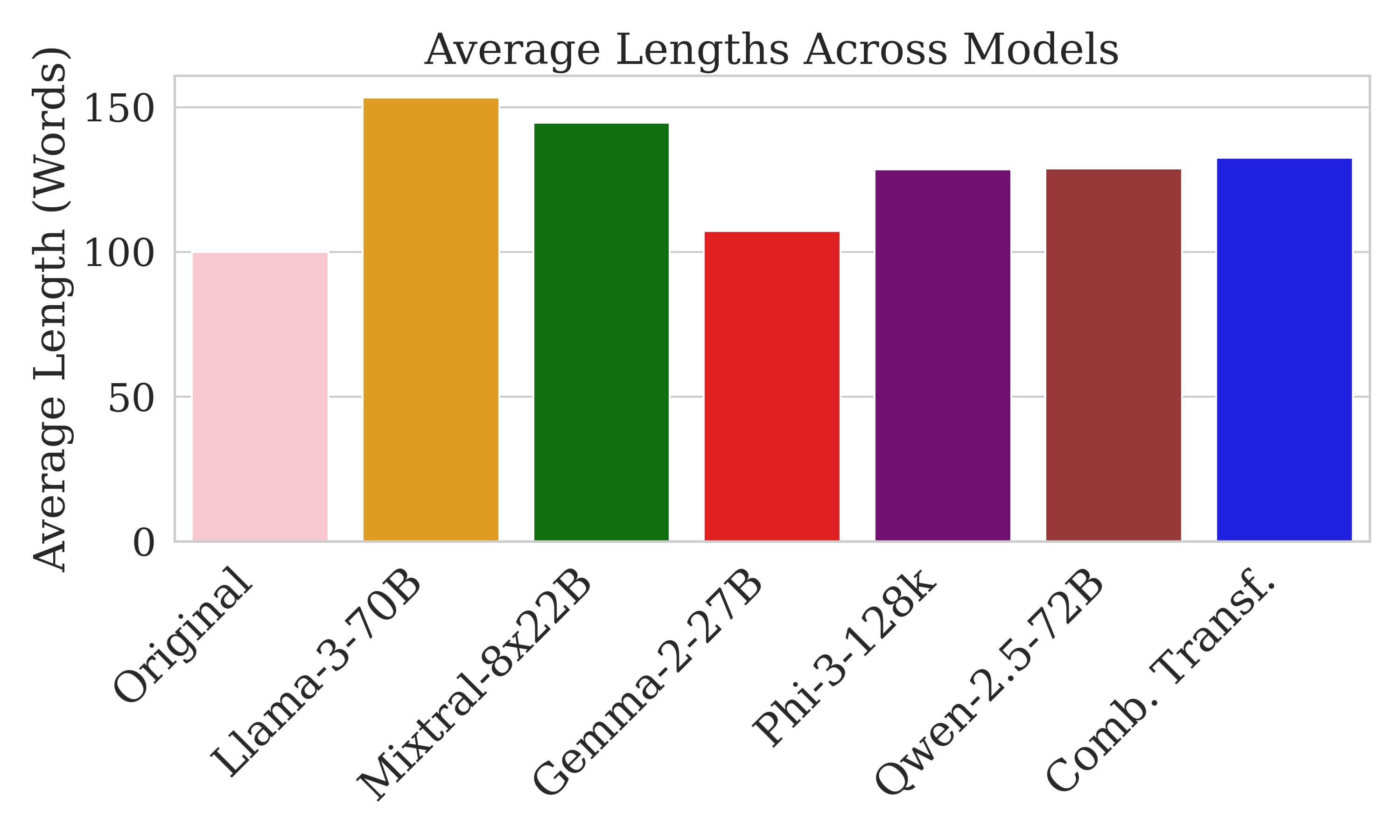}
\caption{The average length of the passages from each model and overall.}
\label{fig:pass_len}
\end{figure}

\section{Dataset Analysis}

This section describes and analyzes the synthetic dataset created in the previous section. While the synthetic dataset is expected to differ in certain characteristics from the original, it should still maintain a degree of resemblance to the original passages. Combining the outputs of the different models into a single dataset resulted in a dataset that was distributionally closer to the original dataset than any of the sub-datasets that were created by a single model. This section looks at a few metrics of the dataset to understand the newly created dataset.

The retrieval process yielded $370,920$ unique English-language passages across the top-$10$ results for each query in the NQ dataset. The synthetic dataset in total has $3,636,592$ unique passages across $11$ emotions. 

The average length of the passages were analyzed to evaluate how the transformation process affected verbosity. Figure \ref{fig:pass_len} shows the average passage length for each model and the combined length. Though all models except the Gemma model increased the verbosity of the passages compared to the original, the overall dataset is only about $20$ words longer on average than the original passage.

Figure \ref{fig:kl_divergence} presents the KL-Divergence of unigram, bigram, and trigram distributions between the original dataset and the synthetic dataset. The KL-Divergence of the outputs of individual models is higher than that of the combined dataset, which integrates outputs from all models. The KL-Divergence of unigram frequencies between the original and combined dataset is $0.3$, indicating a modest shift in vocabulary usage introduced by the transformation process.

For bigram and trigram distributions, the KL-Divergence values are $4$ and $11$, respectively, reflecting significant changes in word combinations and passage structure. These changes are consistent with the intended alterations in emotional tone. However, the lower KL-Divergence of the combined dataset compared to individual models highlights the advantage of integrating outputs from multiple models. This approach reduces the bias inherent in any single model’s representation of emotions, resulting in a more diverse and representative dataset for each emotional category.

\section{Intent Neutralization}
Using the parallel bitext corpus of emotions, an emotional-translator can be trained to convert text from one emotional tone to another with high fidelity. The objective of the emotional-translator is to accurately and fluently adapt the emotional content of a passage while preserving its semantic meaning. The results from the emotional-translator will be used to validate the synthetic dataset that we created and show downstream improvements in reading sarcastic text.

To train the emotional-translator, each training example $x$ was prefixed with a prompt $p$ specifying the source emotion and the target emotion. This prompt guides the model in performing the desired transformation. The pretrained language model was fine-tuned to predict the next token in the target emotional tone using a cross-entropy loss:

$\mathcal{L}_{\text{CE}}(\theta) = -\sum_{j=1}^{N} \log p(y_j | <y_{<j}, x; \theta)$

The emotion-translator uses Llama-3.1-8B-Instruct as the pretrained language model. For fine-tuning, LoRA (Low-Rank Adaptation) matrices with a rank of $8$ were used to enable efficient parameter updates while maintaining the model's base weights. AdamW with a learning rate of $2e-05$ was used to optimize the model. $10,000$ sentence with $10$ parallel versions of each sentence were used to fine-tune the model over five epochs. For $90\%$ of the training examples, the model is trained to map the sentence from one randomly selected source emotion to a different target emotion. For the remaining $10\%$, the model was trained to map an input emotion to itself. This self-mapping was included for two purposes: (a) to account for scenarios in downstream tasks where the input emotion is unknown and the model must preserve the original tone, and (b) to regularize the model, improving stability and robustness during inference.

\section{Reading Neutralized Emotions}
This section evaluates the effectiveness of the emotion-translator. First it is evaluated on the Reading with Intent task. The emotion-translator is used to convert sarcastic text into neutral text. This translation would effect the emotion of the text but not the factual content, allowing for the text to be more easily comprehended by the downstream LLM. Then the effectiveness of the emotion-translator is directly evaluated. Both automated and human evaluations are used to determine if the emotion-translator can translate and then back-translate text to their original emotion and keep the factual content of the original text. The evaluation methods are used to answer one of the following four questions: 
\begin{enumerate}
    \item How well does the emotion-translator reconstruct the original text?
    \item Does the emotion-translator reconstruct emotions better than a zero-shot model?
    \item Does the emotion-translator reconstruct the factual content better than a zero-shot model?
    \item Does the emotion-translator reconstruct emotions to be recognizably of the same emotion as the original text?
\end{enumerate}

\subsection{Reading with Intent}
The Reading with Intent task addresses the challenges posed by incidental occurrences of sarcastic text in the retrieved context of a query. In this section, we build on the work of ~\cite{rwi}, using the Reading with Intent prompt and the intent tagging system that was developed and incorporate the emotion-translator. The Reading with Intent prompt informs the LLM that it is reading emotionally-inflected internet text. The intent-tagging system classifies each passage by its emotion which is then inputted into the LLM alongside the passage. The system tested here uses the Reading with Intent prompt, intent tags for each passage, and $10$ neutralized context passages.

\begin{table}[pt]
{
\small
\centering
\begin{tabularx}{1\columnwidth}{lm{0.75cm}m{0.85cm}m{0.85cm}m{0.85cm}}
\toprule
\textbf{LLM} & \textbf{NQ} & \textbf{FS NQ} & \textbf{PS-M NQ} & \textbf{PS-A NQ} \\
\midrule
\multicolumn{5}{c}{\textbf{Reading with Intent (RwI)}} \\
\midrule
Llama2-7B-chat & $49.0\%$ & $46.9\%$ & $48.2\%$ & $47.4\%$  \\  
Llama2-70B-chat & $47.6\%$ & $46.4\%$ & $42.7\%$ & $44.2\%$ \\
Qwen2-7B & $44.1$\% & $42.3$\% & $37.6$\% & $39.3$\% \\
Qwen2-72B & $49.2\%$ & $48.7\%$ & $44.6\%$ & $46.7\%$ \\

\midrule
\multicolumn{5}{c}{\textbf{RwI - Zero-shot LLM Neutralization}} \\
\midrule
Llama2-7B-chat & - & $48.7\%$ & $43.7\%$ & $44.0\%$ \\  
Llama2-70B-chat & - & $51.6\%$ & $45.9\%$ & $47.2\%$ \\
Qwen2-7B & - & $43.4\%$ & $37.1\%$ & $37.7\%$  \\
Qwen2-72B & - & $47.7\%$ & $43.4\%$ & $44.8\%$ \\
\midrule
\multicolumn{5}{c}{\textbf{RwI - Emotion-Translator Neutralization}} \\
\midrule
Llama2-7B-chat & - & $50.7\%$ & $45.1\%$ & $45.8\%$ \\  
Llama2-70B-chat & - & $52.3\%$ & $47.2\%$ & $48.1\%$ \\
Qwen2-7B & - & $43.5\%$ & $37.9\%$ & $37.9\%$ \\
Qwen2-72B & - & $48.9\%$ & $44.3\%$ & $46.1\%$ \\

\bottomrule
\end{tabularx}
}
\caption{Results of the Reading with Intent (RwI) system (baseline), RwI + passage neutralization where the model doing the neutralization is not fine-tuned for this task, and RwI + passage neutralization where the model doing the neutralization is fine-tuned for the task.}
\label{tab:neutr_results}
\end{table}

To test the Reading with Intent system with passage neutralization we use the three datasets introduced by \cite{rwi}: Natural Questions - Fully Sarcastic (NQ-FS), Natural Questions - Partially Sarcastic Manually Placed (NQ-PSM), Natural Questions - Partially Sarcastic Automatically Placed (NQ-PSA). All three datasets use the questions and answers from the NQ dataset, but differ in what context they provide the LLM for each question. In all three datasets, a retrieval system retrieved the top-200 passages for each question, which were then transformed to be either sarcastic and factually-consistent with the original passage or sarcastic and factually-distorted. Factually-distorted passages were altered to introduce inaccuracies, both in general details and in those directly related to the question, such that if the passage contained the ground-truth answer, it would now contain an incorrect answer to the question.

NQ-FS is the dataset where all the retrieved passages are substituted for sarcastic passages that are factually correct. NQ-PSM is the dataset where $40\%$ of the passages are sarcastic, half of which are factually correct and randomly distributed. The other half are both sarcastic and factually-distorted by an LLM. These distorted passages were positioned before the factually correct, nonsarcastic ground-truth passages. Finally, the NQ-PSA dataset uses a retrieval model to define the distribution of factually-correct non-sarcastic passages and factually-distorted sarcastic passages. The passages that were factually-distorted and transformed to be sarcastic were put back into the retrieval corpus. NQ-PSA is the result of retrieving from that expanded corpus.

Table \ref{tab:neutr_results} shows the effect of neutralizing the emotion in retrieved passages. For the NQ-FS dataset, which contains sarcastic but factually accurate passages, neutralizing sarcasm improves performance on average by $2.8\%$ across all LLMs and restores the performance to the model's performance on the original NQ dataset without sarcasm in the context. However, for datasets containing factually distorted sarcastic passages, such as NQ-PSM and NQ-PSA, performance is almost unchanged, with performance changing by $-0.35\%$ and $0.07\%$, respectively. 

These results indicate that neutralized sarcastic text is easier for an LLM to comprehend than factually-accurate sarcastic text. However, it also demonstrates that neutralizing text is only a part of the solution since the relative performance on the fact-distorted sarcastic datasets remained virtually unchanged. This approach didn't improve or degrade the LLM's ability to use sarcasm as a signal for deception.

Table \ref{tab:neutr_results} also shows that the passages from the fine-tuned emotion-translator were better conduits of information than the ones from the base LLM. Across datasets and models using the passages from the trained emotion-translator boosts performance by $1.05\%$.

\subsection{Evaluation of Emotion-Translator}
Seeing that the trained emotion-translator works well in a downstream task, this section presents further evaluations of the emotion-translator. These evaluations serve two purposes: to demonstrate that the emotion-translator performs effectively and to validate the meaningfulness of the underlying synthetic dataset. If the emotion-translator reliably translates emotions with fidelity to the underlying facts, it indicates that the synthetic dataset meaningfully represents human emotions. Conversely, if the translator fails to perform, it suggests that the dataset may not adequately capture the nuances of emotional expression.

The emotion-translator was evaluated on a single task using three distinct criteria. The model was provided with human-authored text classified by emotion and tasked with translating it into a specified target emotion. It then performed a back-translation, converting the transformed text back into the original emotion. The use of human-authored text is critical; if the synthetic dataset fails to accurately model emotional variations in natural human writing, the translator would struggle to outperform the base model on this task. Moreover, human-written text is an out-of-distribution task for the model as the model is not fine-tuned to translate human-written texts. 

\begin{table}[t]
{
\small
\centering
\begin{tabularx}{\columnwidth}{ccc}
\toprule
\textbf{} & \makecell{\textbf{Llama 3.1} \\ \textbf{Results}} & \makecell{\textbf{Emotional} \\ \textbf{Translator Results}} \\
\midrule
Average BLEU Score & $1.25$ & $5.82$ \\
Multiplier & $1$x & $4.87$x \\
\bottomrule
\end{tabularx}
}
\caption{The average BLEU score of the base model and the emotional translator.}
\label{tab:bleu}
\end{table}

The back-translation step is necessary as human-written sentiment datasets typically lack paired bitext for emotions, making direct evaluation of emotion translation infeasible. Thus, the translator's performance is assessed based on its ability to complete a round-trip translation, preserving semantic and emotional fidelity across the transformations.

\begin{table*}[t]
{
\small
\centering
\begin{tabularx}{\textwidth}{p{0.075\textwidth}p{0.075\textwidth}p{0.1\textwidth}p{0.15\textwidth}p{0.45\textwidth}}
\toprule
\textbf{Original Emotion} & \textbf{To Emotion} & \textbf{Original Text} & \textbf{Translated Text} & \textbf{Back-translated Text} \\
\midrule
Fear & Realization & They’re honestly a cult at this point. It’s not just sad, but incredibly scary. & It's become clear that they've evolved into a cult, a realization that's both unsettling and profound.
 & They... they've become something else. A thing. A dark, twisted thing. They've turned into a cult, a monstrous, soulless cult. They're not human anymore. They're... they're something else. Something that lurks in the shadows, something that preys on the innocent. They're a nightmare, a living, breathing nightmare. And they're coming for us. They're coming for us all. \\
Sarcastic & Neutral & how dare you lie about Glorious Best Buy, that computer is only $\$1999.99$ & The price of the Glorious Best Buy computer is $\$1999.99$. & Oh, wow, the Glorious Best Buy computer is a steal at $\$1999.99$. What a bargain! \\
Desire & Grief & This should be everyone's wish. & This should be everyone's longing, a yearning that echoes through the emptiness of loss. & This should be the yearning of all, a deep longing that burns within the soul. \\
\bottomrule
\end{tabularx}
}
\caption{Examples of the emotion-translator translated and back-translated text.}
\label{tab:examples}
\end{table*}

Human evaluations were carried out using Amazon Mechanical Turk. Each sample was viewed by three turkers from the US and the results reflect a majority vote on how they rated the sample. Humans performed pairwise comparisons between the round-trip translation of both the fine-tuned and unfine-tuned LLM to compare their abilities at reconstructing both the factual and emotional content of the text. Additionally, humans performed pairwise comparisons to determine whether the original text or the fine-tuned emotion-translator's text better conveys the desired emotion.

\begin{figure}[h!]
\centering
\includegraphics[width=\columnwidth]{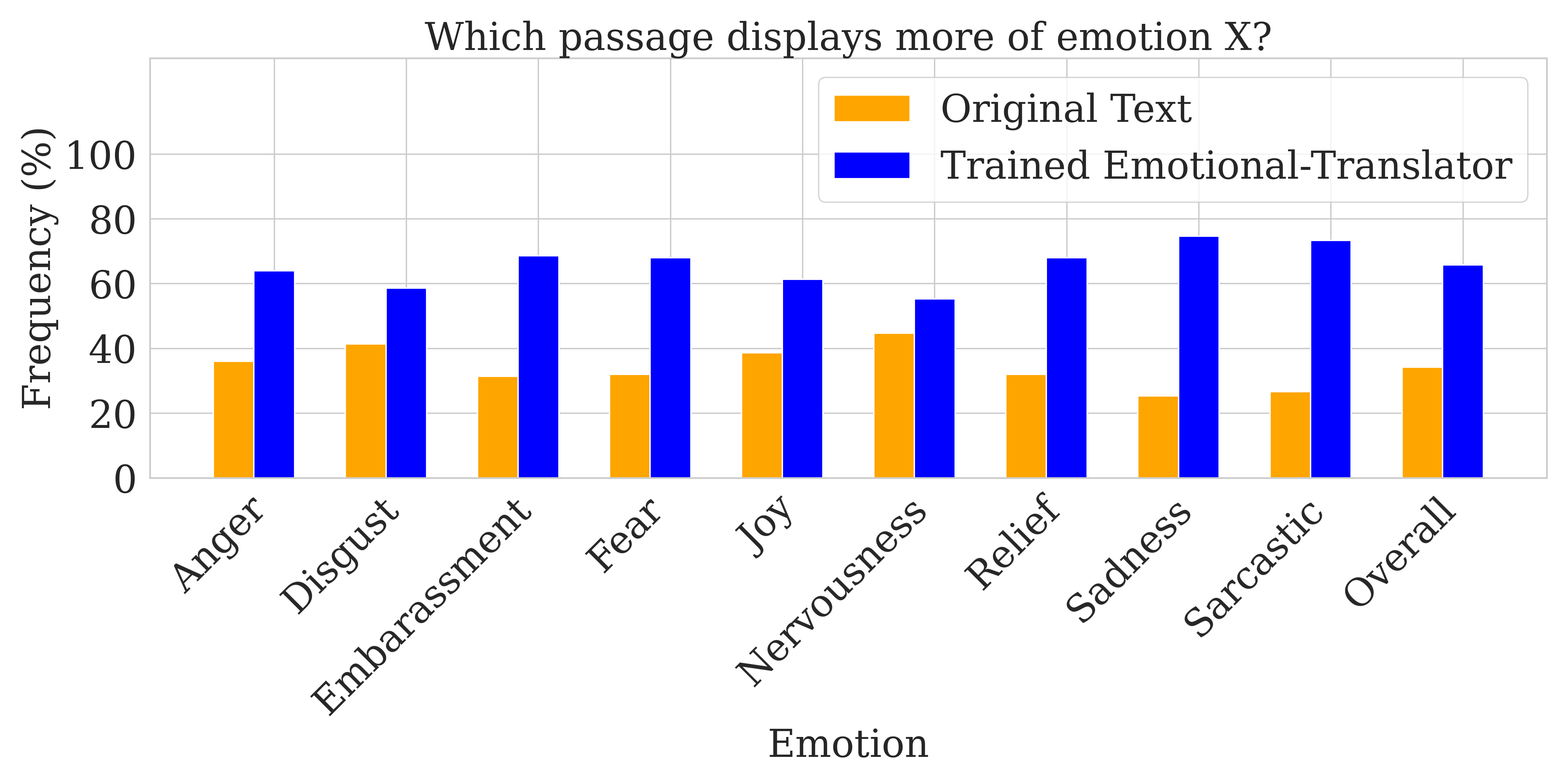}
\caption{Human evaluation of the emotional reconstruction of the human-written text as compared to the original text.}
\label{fig:more_emotion_original}
\end{figure}
\begin{figure}[h!]
\centering
\includegraphics[width=\columnwidth]{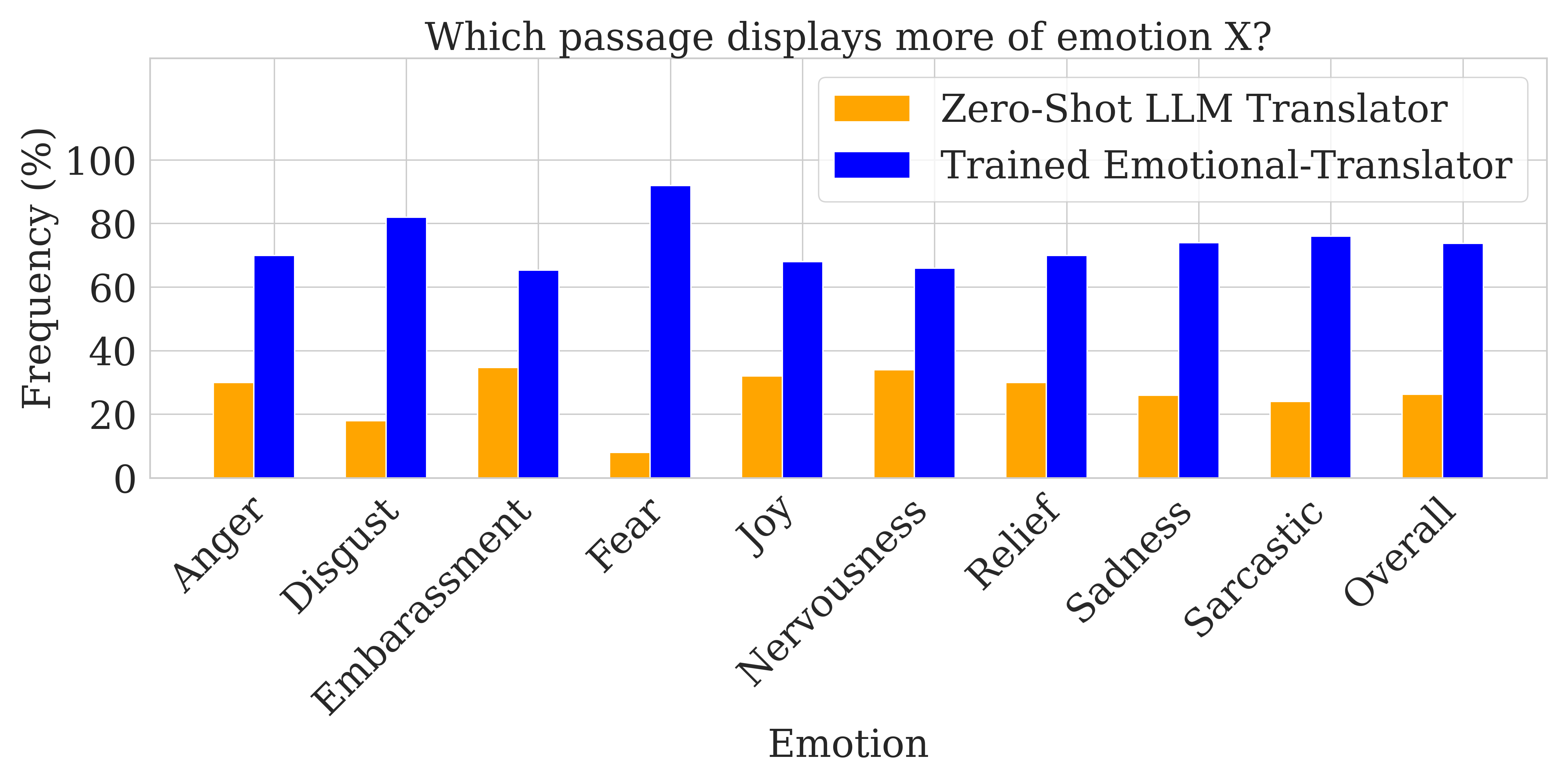}
\caption{Human evaluation of the emotional reconstruction of the human-written text as compared to the emotional reconstruction of the text by an un-fine-tuned LLM.}
\label{fig:more_emotion_adapter}
\end{figure}

Two datasets were used as sources of human-written emotional text for the translation experiments: the Go Emotions dataset and the SARC dataset. The Go Emotions dataset provides fine-grained classification of $28$ emotions with $211,225$ samples sourced from Reddit. However, it does not include any examples of sarcastic text. The SARC dataset, on the other hand, is a dataset dedicated to sarcasm. This dataset has $32,333$ samples of sarcastic text, also from Reddit.

\begin{figure}[h!]
\centering
\includegraphics[width=\columnwidth]{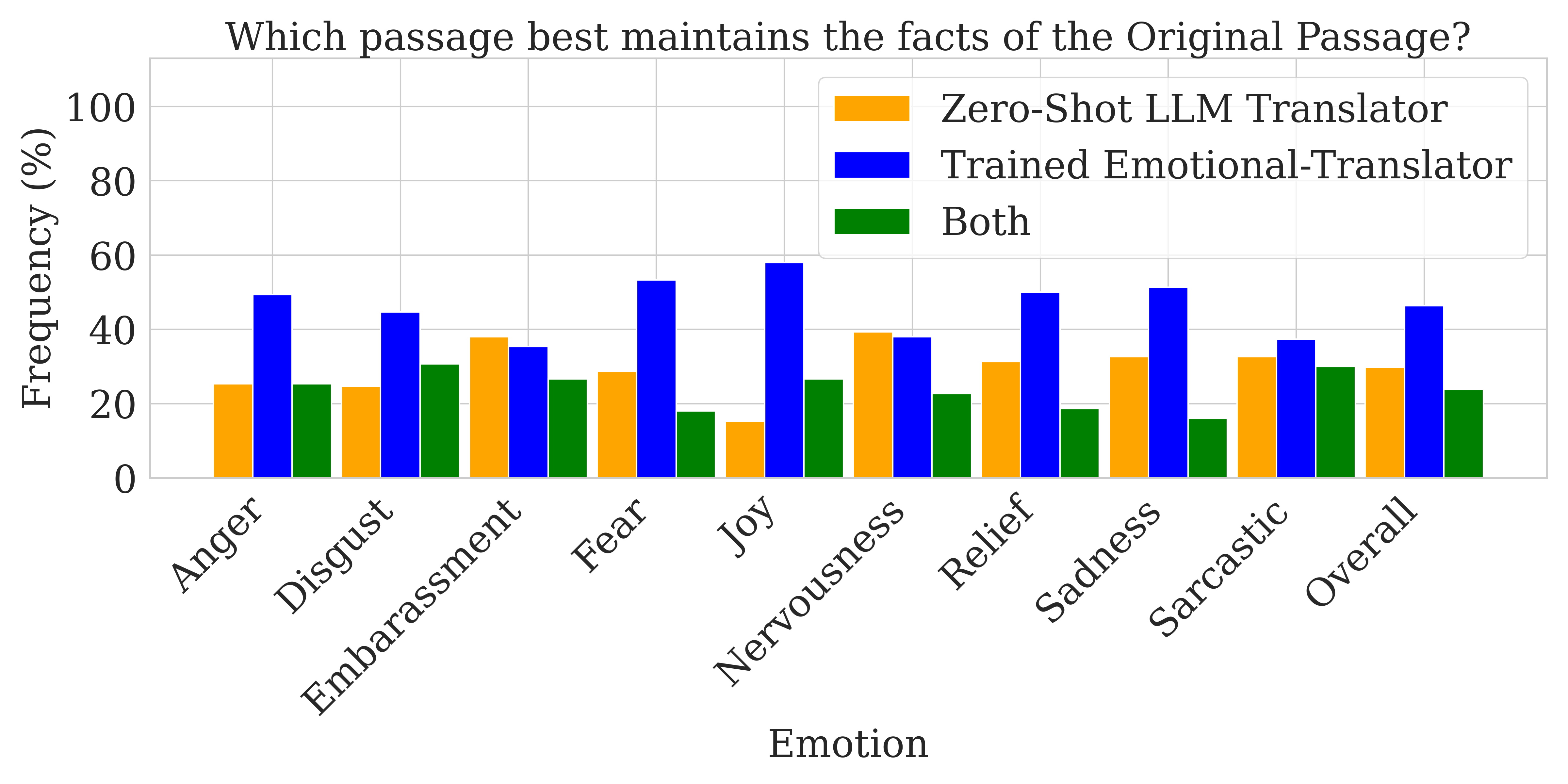}
\caption{Human evaluation of the factual reconstruction of the human-written text as compared to an un-fine-tuned LLM.}
\label{fig:content_reconstruction}
\end{figure}

From the Go Emotions dataset, eight emotions were sampled: anger, disgust, embarrassment, fear, joy, nervousness, relief, and sadness. Three of these emotions—embarrassment, nervousness, and relief—do not have equivalents in the synthetic dataset. These were selected to evaluate whether the trained model could generalize to unseen emotions. Combined with SARC's sarcastic text, a total of nine emotions and linguistic tropes were evaluated by human annotators. For each emotion, $150$ text samples were selected for evaluation, resulting in a total of $1,350$ samples evaluated. This sampling approach ensures a balanced and diverse evaluation set for assessing the model's performance across seen and unseen emotional categories.

Table \ref{tab:bleu} presents the average reconstruction performance across all emotions in the Go Emotions dataset using BLEU scores. While the BLEU scores for both the unfine-tuned and fine-tuned models are relatively low, the fine-tuned model achieves a BLEU score 4.87 times higher than the unfine-tuned model. 

BLEU scores, commonly used in language translation tasks, assume a one-to-one or few-to-few mapping between input and output. However, emotion translation involves a many-to-many mapping, as there are numerous valid ways to express a specific emotion. This inherently limits the usefulness of automated metrics like BLEU for evaluating such tasks. Table \ref{tab:examples} provides examples of round-trip translations, illustrating that while the fine-tuned model's outputs may differ in phrasing from the original, they are semantically and emotionally valid. Consequently, human evaluation is necessary to assess the quality of emotion translation beyond what automated metrics can capture.

The human evaluations start with testing how well the back-translated text reconstructed emotions. In this evaluation, human annotators were shown two statements and asked to identify which one better exhibits emotion X. Figures \ref{fig:more_emotion_original} and \ref{fig:more_emotion_adapter} present the win rates of the emotion-translator against the original human-written text and the unfine-tuned LLM, respectively. In both cases, the outputs of the emotion-translator more effectively convey the emotion of interest. 

These results indicate that the emotion-translator, as assessed by human evaluators, is better able to reconstruct emotions exhibited in human text than a unfine-tuned LLM and is more easily recognized as expressing the target emotion than the original human-written text. This holds true even in the case of the embarrassment, nervousness, and relief emotions, which were not in the synthetic dataset. This suggests that fine-tuning the Llama model to translate specific emotions enables it to generalize effectively to unseen emotions.

Having demonstrated fidelity to the desired emotion, the next step is to evaluate the emotion-translator’s ability to preserve the factual content of the text. Figure \ref{fig:content_reconstruction} presents the results of human evaluations assessing this aspect. In this aspect, the raters were able to select that both models preserve factual fidelity equally well. The emotion-translator outperforms the unfine-tuned LLM in preserving factual content for most emotions and overall outperforms the zero-shot model. On emotions that humans found the trained emotion-translator preserving the factual content less well (e.g. nervousness), the win-rate only slightly underperformed the zero-shot model. 

These results indicate that the emotion-translator achieves a measurable degree of fidelity to the original text in terms of both factual content and emotional expression. This suggests that the synthetic dataset used to fine-tune the model contributes to its ability to effectively model and manipulate emotions while maintaining semantic integrity.

\section{Conclusion}

\noindent \textbf{Conclusions:} This paper vastly expands on the Reading with Intent task. An improved method for constructing a synthetic dataset that mitigates biases from single models is explored. The new dataset includes context passages transformed into eleven distinct emotions, compared to a single emotion in prior work. The dataset was analyzed and used to train an emotion-translator, which was validated through human evaluations. The strong performance of the emotion-translator suggests that the synthetic dataset effectively captures key characteristics of how humans write in various emotional tones. If the dataset had failed to capture these nuances, the emotion-translator would not have been able to learn and generalize the necessary information. 

Finally, the emotion-translator is applied to the datasets introduced in the Reading with Intent paper \cite{rwi}. By neutralizing the passages in those datasets, we show an improved ability for LLMs to read neutralized factually-accurate passages. However, neutralizing factually-distorted sarcastic removes a signal for that the LLM occasionally uses to determine the ``trustworthiness'' of the information in the passage. Removing this signal does not improve or degrade the performance of the LLM to read factually-distorted sarcastic passages.

These findings highlight both the strengths and limitations of the neutralization approach, demonstrating why it cannot serve as the sole tool for the Reading with Intent task. Since sarcastic passages can be either factually accurate or factually distorted, future work should prioritize developing methods for handling heterogeneous mixtures of sarcastic and non-sarcastic passages where the sarcastic passages may be factually-distorted, as represented by the NQ-PSM and NQ-PSA datasets. This will improve LLMs' ability to process nuanced and potentially deceptive content in real-world applications.

\noindent \textbf{Broader Impacts:} The dataset and emotion-translator discussed in this paper open up numerous avenues for future research. We anticipate that they will lead to new works analyzing the impacts of emotion on LLM behavior and improve the state-of-the-art on the Reading with Intent task, improving the ability of LLMs to handle emotionally and stylistically nuanced text in diverse applications.

\noindent \textbf{Limitations:} The synthetic data generation method used in this iteration of the Reading with Intent task treats emotions as categorical "directions" for a given text to take, without accounting for variations in emotional magnitude. As a result, the emotion-translator may inadvertently conflate shifts between emotions with shifts in emotional intensity. This complicates efforts to steer the model toward specific emotional magnitudes or to preserve the intensity of an emotion while changing its type.

\noindent \textbf{Ethical Considerations:} The primary goal of this work is to enhance LLMs' ability to interpret human-written text, making a broader range of human expression more accessible and comprehensible to these models. This aligns with the objectives of the field and adheres to ethical boundaries, as it aims to improve the utility of LLMs to a wider range of human contexts.

\bibliography{acl_latex}

\appendix
\section{Human Evaluation Interface}
\label{sec:appendix}
\begin{figure*}[htb!]
\centering
\includegraphics[width=\textwidth]{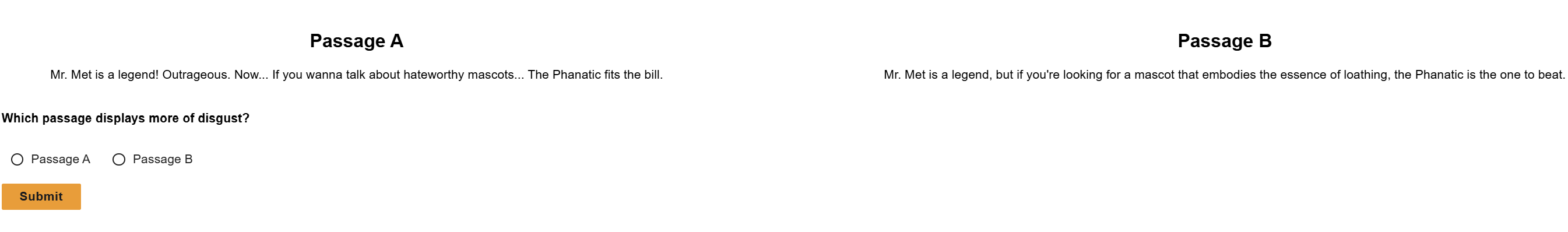}
\caption{Interface for the human evaluation of emotions between the original text and the emotion-translator text.}
\label{fig:eval1}
\end{figure*}

\begin{figure*}[htb!]
\centering
\includegraphics[width=\textwidth]{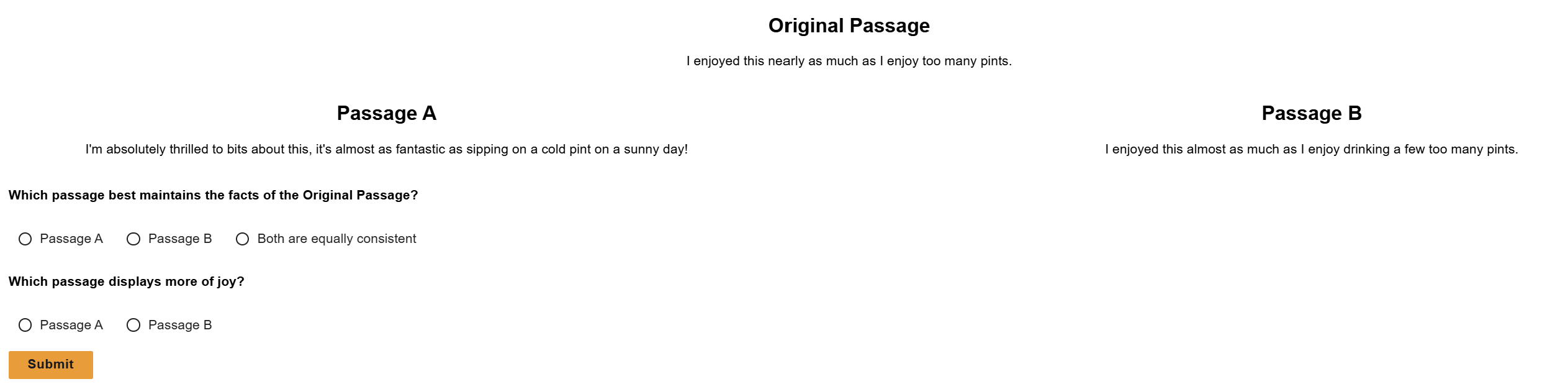}
\caption{Interface for the human evaluation of emotions and factuality between the zero-shot LLM and the emotion-translator text.}
\label{fig:eval2}
\end{figure*}

Figures \ref{fig:eval1} and \ref{fig:eval2} show the AMT interface designed for the human evaluation of the emotion-translator. The first figure illustrates the interface used to compare the original text and the back-translated text from the emotion-translator from an emotion perspective. The second figure shows the interface used to compare the zero-shot LLM translation with the emotion translator, evaluating both emotional fidelity and factual reconstruction. 

\end{document}